\documentclass[runningheads]{llncs}
\usepackage[T1]{fontenc}
\usepackage{graphicx}
\usepackage{amssymb}
\usepackage{amsmath}
\usepackage{subfigure}
\usepackage{booktabs}

\usepackage[pagebackref=true,breaklinks=true,letterpaper=true,colorlinks,citecolor=blue,linkcolor=blue,bookmarks=false]{hyperref}
\usepackage{multirow}

\begin{document}
\title{Polyp-DAM: Polyp segmentation via depth anything model}
%
%
\author{Zhuoran Zheng\inst{1} \and
Chen Wu\inst{2} \and
Wei Wang\inst{3} \and 
Yeying Jin\inst{4} \and 
Xiuyi Jia\inst{1}
}
%
%
\institute{Nanjing University of Science and Technology \and
University of Science and Technology of China \and Sun Yat-sen University \and National University of Singapore \\
\email{zhengzr@njust.edu.cn}}
\maketitle              
\begin{abstract}
%
%
Recently, large models (Segment Anything model) came on the scene to provide a new baseline for polyp segmentation tasks.
This demonstrates that large models with a sufficient image level prior can achieve promising performance on a given task.
In this paper, we unfold a new perspective on polyp segmentation modeling by leveraging the Depth Anything Model (DAM) to provide depth prior to polyp segmentation models.
Specifically, the input polyp image is first passed through a frozen DAM to generate a depth map.
The depth map and the input polyp images are then concatenated and fed into a convolutional neural network with multiscale to generate segmented images.
Extensive experimental results demonstrate the effectiveness of our method, and in addition, we observe that our method still performs well on images of polyps with noise.
The URL of our code is \url{https://github.com/zzr-idam/Polyp-DAM}.

\end{abstract}
\section{Introduction}
The appearance of polyps in the intestine is a precursor to rectal cancer~\cite{shussman2014colorectal}.
Therefore, the detection of polyps is of great importance for the early diagnosis and treatment of rectal cancer.
However, manual screening by physicians usually leads to the problem of underdetection and misdetection. So far, automated polyp segmentation methods provide physicians with accurate decision-making.

\begin{figure}[t]
	\begin{center}\scriptsize
		\tabcolsep 1pt
		\begin{tabular}{@{}c@{}}
                \includegraphics[width = 0.45\textwidth]{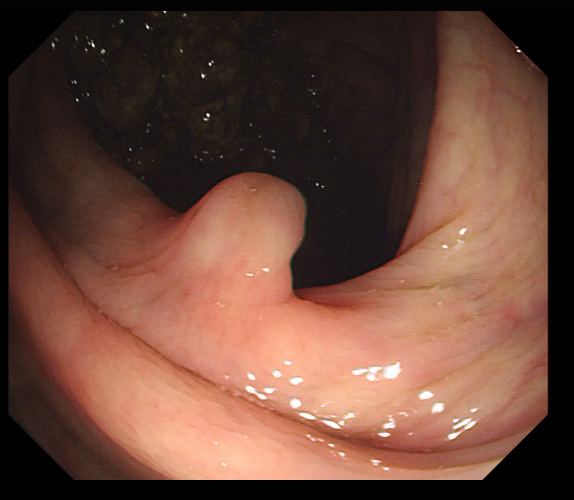}  
                \includegraphics[width = 0.45\textwidth]{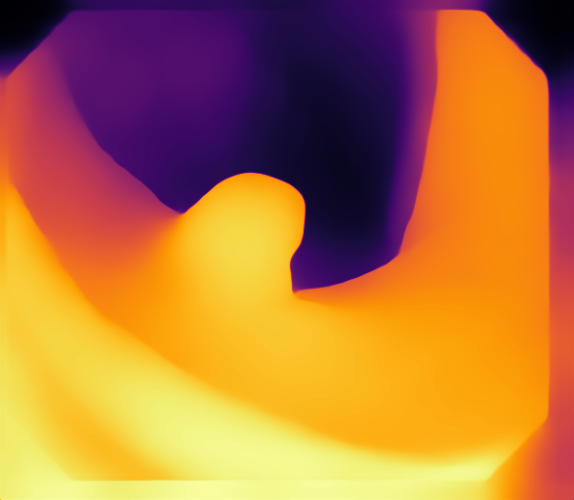}  
		\end{tabular}
	\end{center}
 \vspace{-2mm}
 	\caption{This figure shows the result of processing a polyp image using DAM, which distinguishes the polyp (foreground) from the rest of the image (background) very well.}
	\label{f1}
 \vspace{-2mm}
\end{figure}

Currently, several automated polyp segmentation methods based on deep learning have been developed and made significant progress~\cite{zhao2021automatic,tajbakhsh2015automated,duc2022colonformer,wang2022stepwise,zhou2023cross,biswas2023polyp,li2023polyp,zhou2023can}.
In particular, the large model-based approach~\cite{biswas2023polyp,li2023polyp,zhou2023can} (SAM) provides a new baseline for polyp segmentation.
The outstanding performance of these methods on the polyp problem benefits from the role of large models that carry a large number of prior image levels.
However, SAM-based methods require large computational resources to fine-tune polyp segmentation datasets, and the fine-tuning results are fraught with uncertainty.
To address this problem, we use the image-level prior (see Figure~\ref{f1}) provided by the Depth Anything Model (DAM)~\cite{yang2024depth,huang2024endo} to explicitly construct a polyp segmentation model, which does not require fine-tuning of the larger model.

\begin{figure}[t]
	\begin{center}\scriptsize
		\tabcolsep 1pt
		\begin{tabular}{@{}c@{}}
                \includegraphics[width = 0.95\textwidth]{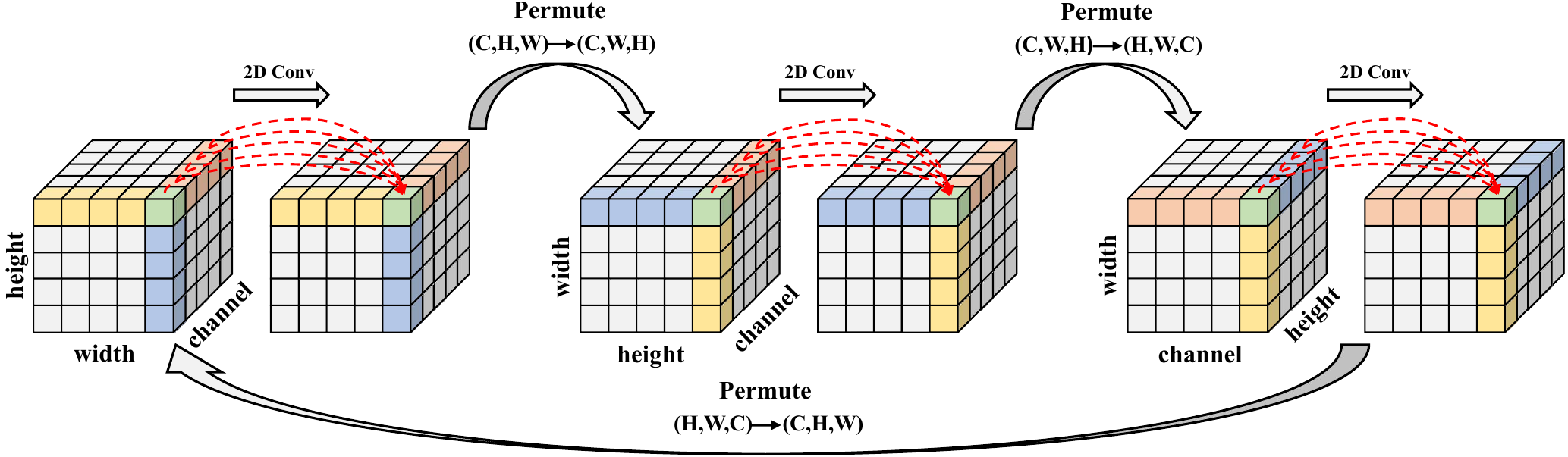}  
		\end{tabular}
	\end{center}
 \vspace{-2mm}
 	\caption{This figure shows the process of processing our global module. Here, the transformation of the feature map dimension is followed by filtering using a $1 \times 1$ convolution, similar to performing an attention operation on this dimension.}
	\label{f2}
 \vspace{-4mm}
\end{figure}

In this paper, we propose a simple but effective polyp segmentation method, named Polyp-DAM.
Specifically, first, the input polyp image is performed by the DAM to obtain a depth image corresponding to the polyp.
Depth images and polyp images are sampled to four different sizes and fed into our designed multi-scale MixNet (M$^{2}$ixNet).
M$^{2}$ixNet is inspired by the vanilla MLP-Mixer~\cite{MlpMixer}, but different from it, the global module encodes the feature maps of images from the perspectives of the width $W$, height $H$, and channel $C$ (see Figure~\ref{f2}).
Through dimensional transformation operations ($\texttt{permute}$), the information encoded in each feature map is associated and fused. Note that dimension transformation operations do not introduce additional parameters, which means that by permuting the feature maps from different views, the global module achieves efficient modeling of long-range dependency.
Furthermore, to efficiently model the local features of the feature map, we design a local feature capture module.
The local module filters the feature maps produced by the global module through a 3D convolution for local modeling (here the feature map is viewed as a grid).
Finally, the four feature maps output by M$^{2}$ixNet with different scales are fused by a 2D convolution to obtain a mask corresponding to a polyp image.

\begin{itemize}
\item To the best of our knowledge, we are the first to introduce the DAM prior to the polyp image segmentation task. The DAM can assist the segmentation model in distinguishing between foreground and background.

\item We propose M$^{2}$ixNet, which efficiently performs the segmentation of polyp images through global versus local modeling. It is worth noting that the number of parameters in our network is only \textbf{0.47M} without considering DAM.

\item Extensive experiments demonstrate that our method outperforms state-of-the-art approaches with better quantitative and qualitative evaluation.
\end{itemize}

\section{Related Work}
%
%
%
UNet~\cite{ronneberger2015u} is a classical image segmentation model with a symmetric encoding-decoding architecture, and it can efficiently fuse shallow and deep features of an image.
Based on its good segmentation performance, some polyp segmentation methods~\cite{yin2022duplex,patel2021enhanced,wu2021precise,poudel2021deep} adopt the U-shape network.
In addition, some methods use attention mechanisms and carefully designed strategies to improve segmentation accuracy~\cite{fan2020pranet,wei2021shallow,zhang2020adaptive,zhou2023cross,wang2022ffcnet,yu2022frequency,chi2020fast,suvorov2022resolution}.
Recently, since the success of the Transformer in multiple fields, some approaches~\cite{dong2021polyp,wang2022stepwise,duc2022colonformer,biswas2023polyp,li2023polyp,zhou2023can} have replaced the backbone of networks from CNN with Transformer and obtained good results.
In this paper, we focus on obtaining accurate polyp image segmentation results using an image-level prior.

\begin{figure}[t]
	\begin{center}\scriptsize
		\tabcolsep 1pt
		\begin{tabular}{@{}c@{}}
                \includegraphics[width = 1.0\textwidth]{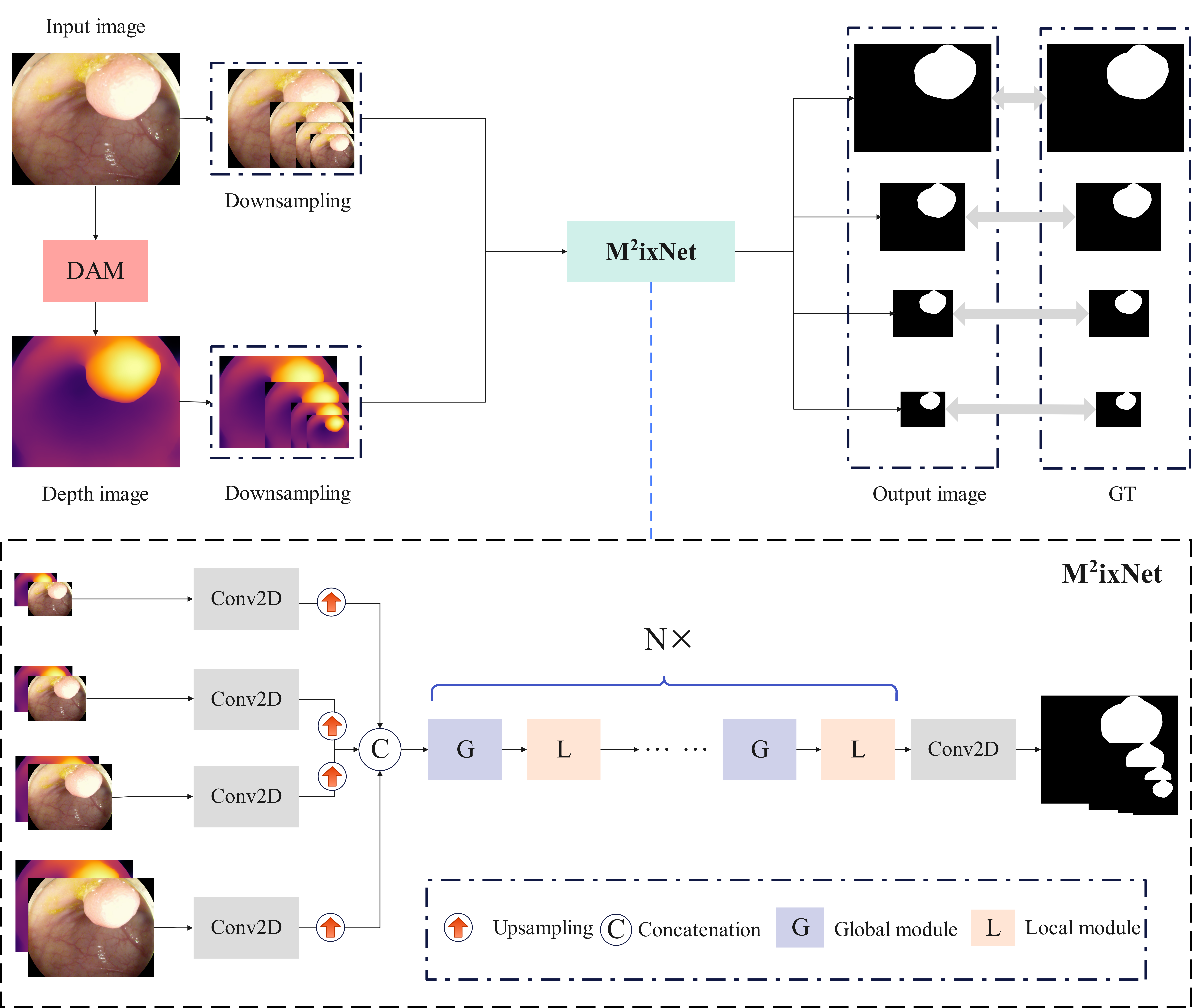}  			
		\end{tabular}
	\end{center}
 \vspace{-2mm}
 	\caption{The structure of our method. First, we obtain the depth map of the input polyp image via DAM. Next, the input image and depth map are bilinearly downsampled into $256 \times 256$, $128 \times 128$, $64 \times 64$, and the original resolution is fed into M$^{2}$ixNet. Finally, our network outputs four masks of different sizes and four sizes generated on GT for learning. It is worth noting that our network used in the evaluation on the benchmark is the image at the model output original resolution.}
	\label{overview}
 \vspace{-4mm}
\end{figure}

\section{Method}
An overview of Polyp-DAM is shown in Fig.~\ref{overview}. For a given polyp image $\mathbf{X}\in\mathbb{R}^{H\times W\times3}$, we start by projecting $\mathbf{X}$ onto a 4-level resolution image \{$\mathbf{X}_{o},\mathbf{X}_{64},\mathbf{X}_{128}, \\  \mathbf{X}_{256}$\} by using the bilinear interpolation.
Then, we input $\mathbf{X}$ into the DAM to generate a depth map $\mathbf{D}\in\mathbb{R}^{H\times W\times3}$, which is also projected onto the image at 4 levels of resolution \{$\mathbf{D}_{o},\mathbf{D}_{64},\mathbf{D}_{128},\mathbf{D}_{256}$\}.
Finally, these depth maps of different resolutions with corresponding polyp images are input into the M$^{2}$ixNet to get an accurate mask $\mathbf{M}$.
Here, different scales of polyp information are input into a convolutional layer ($3 \times 3$ convolution kernel) in pairs respectively, followed by output feature maps up-sampled to the resolution of the original image.
It is worth noting that the accurate mask $\mathbf{M}$ was also mapped to 4 resolutions to do loss in the training stage for the GT $\mathbf{O}$.
%

\subsection{Global Module of M$^{2}$ixNet}
The polyp images at different scales and the corresponding depth images are convolved to obtain a feature map $\mathbf{F}_{0} \in\mathbb{R}^{H\times W\times 12}$ by concatenating them on the channel domain, respectively.
After that, $\mathbf{F}_{0}$ is fed into the global module G initially.
G is designed from~\cite{MlpMixer}. We try to capture the long-range dependence of polyp images using a small number of operators.
Specifically, we first adjust the normalized input features' resolution, then perform some dimension transformation operations on them. Given the input feature $\mathbf{F}_{0}$, this procedure can be written as:
\begin{align}
    \mathbf{F}_{1}&=\text{IP}(\mathbf{F}_{0}),\\
    \hat{\mathbf{F}}_{1}&=\underbrace{\text{Permute}(\text{ReLU}(\text{Conv}_{1\times1}(\mathbf{F}_{1})))}_{\times3},
\end{align}
where $\hat{\mathbf{F}}_{t}$ is intermediate result. $\text{IP}(\cdot)$ corresponds to the interpolation operation, $\text{Conv}_{1\times1}(\cdot)$ is a $1\times1$ convolution, $\text{ReLU}(\cdot)$ represents ReLU function, $\text{Permute}(\cdot)$ denotes the dimension transformation operation and $\times 3$ denotes three operations in sequence. It is worth noting that before executing the last dimension transformation operation, we replace the ReLU function with the Sigmoid function.

Afterward, We use interpolation to adjust the feature $\hat{\mathbf{F}}_{t}$ to its original resolution to estimate the attention map and adaptively modulate the input $\mathbf{F}_{0}$ according to the estimated attention via element-wise product. This process can be written as:
\begin{equation}
    \mathbf{F}_{g}=\text{IP}(\hat{\mathbf{F}}_{1})\odot \mathbf{F}_{0},
\end{equation}
where $\mathbf{F}_{g}$ are the final output features and $\odot$ represents element-wise product.

\subsection{Local Module of M$^{2}$ixNet}
For local feature extraction, we mainly use a 3D convolution to perform on the input feature map.
Specifically, given the inputs $\mathbf{F}_{0}$ and $\mathbf{F}_{g}$, they are expanded by one dimension thus obtaining $\hat{\mathbf{F}}_{0} \in\mathbb{R}^{1 \times H\times W \times C}$ and $\hat{\mathbf{F}}_{g} \in\mathbb{R}^{1 \times H\times W \times C}$.
Next, $\hat{\mathbf{F}}_{0}$ and $\hat{\mathbf{F}}_{g}$ are concatenated in the newly expanded dimensions to obtain a tensor $\mathbf{B} \in\mathbb{R}^{2 \times H\times W \times C}$.
Then, we use a 3D convolution to extract the features of $\mathbf{B}$ to yield a new feature map $\mathbf{F}_{2} \in\mathbb{R}^{H\times W \times C}$(here the input channel of the 3D convolution is 2 and the output channel is 1).
Finally, $\mathbf{F}_{2}$ passes through a convolutional block (including two convolutional layers with $3 \times 3$ convolutional kernels) to obtain a feature map that is summed with $\mathbf{F}_{0}$ performing the corresponding element.
The above process can be written:
\begin{equation}
    \mathbf{F}_{l}= \text{ConvB}(\text{Conv3D}(\text{Cat}(\text{UnSqu}(\mathbf{F}_{0}),\text{UnSqu}(\mathbf{F}_{g})))) + \mathbf{F}_{0},
\end{equation}
where $\mathbf{F}_{l}$ are the final output features, UnSqu denotes the expanded dimension, ConvB denotes the convolution block and $+$ represents the element-wise sum.

In this paper, we use 32 pairs of global and local modules.
Among other things, we use a lot of shortcuts between modules.
At the output of the network, a 2D convolution is used to squeeze the channels of the feature map to 1.

\subsection{Loss Function}
Our loss function can be formulated as :
\begin{equation}
\mathcal{L} = \mathcal{L}_{IoU}^{w} +\mathcal{L}^{w}_{BCE},
\end{equation}
where $\mathcal{L}_{IoU}^{w}$ and $\mathcal{L}^{w}_{BCE}$ represent the weighted intersection over union loss and weighted binary cross entropy loss~\cite{wei2020f3net}, which are adopted for the global and local restriction. Deep supervision is performed on the four stage outputs ($i.e.,$ $\mathbf{M}_{o}$, $\mathbf{M}_{64}$, $\mathbf{M}_{128}$ and $\mathbf{M}_{256}$). Each output is up-sampled to the same size as the ground truth $\mathbf{O} \in \{\mathbf{O}_{o}, \mathbf{O}_{64}, \mathbf{O}_{128}, \mathbf{O}_{256}\}$. The total loss can be formulated as:
\begin{equation}
\mathcal{L}_{total} = \sum\limits_{i=1}^{i=4} \mathcal{L}(\mathbf{M}_{i},\mathbf{O}_{i})
\end{equation}
In this paper, we use the output of $\mathbf{M}_{o}$ as the segmented image generated by the model.
Note that, except for the original image resolution $\mathbf{M}_{o}, \mathbf{O}_{o}$, all other images are obtained by bilinear interpolation.

\section{Experiment}
\subsection{Implementation Details}
We implement our model in the PyTorch 2.0 framework. The GPU shader is NVIDIA TITAN RTX 3090 with 24G RAM. During training, all input images are uniformly resized to $352 \times 352$ and a multi-scale training strategy~\cite{fan2020pranet} is employed. The network is trained for 200 epochs with a batch size of 16. For optimizer, we use the AdamW~\cite{loshchilov2017decoupled} to update parameters. The learning rate is 1e-4 and the weight decay is 1e-4. 

\subsection{Datasets and Metrics}
To evaluate the performance of the proposed M$^{2}$ixNet,  we follow the popular experimental setups~\cite{fan2020pranet}, which adopt 5 publicly available benchmarks:
Kvasir-
SEG~\cite{jha2020kvasir}, CVC-ClinicDB~\cite{bernal2015wm}, ColonDB~\cite{tajbakhsh2015automated}, Endoscene~\cite{vazquez2017benchmark} 
and 
ETIS~\cite{silva2014toward}. 
The training set comprises 900 images from Kvasir-SEG and 550 from CVC-ClinicDB. The remaining 100 images from Kvasir-SEG and 62 images from CVC-ClinicDB are used as test sets. Since they are already seen during training, we use them to assess the learning ability of our model. We perform testing using ColonDB, Endoscene, and ETIS, as these datasets are not used during the training stage, they are utilized to measure the generalization ability of our model.

Following previous work~\cite{fan2020pranet,dong2021polyp}, We employ six widely-used metrics for quantitative evaluation, including Mean Dice Similarity Coefficient (mDice), mean Intersection over Union (mIoU),  Mean Absolute Error (MAE),  weighted F-measure ($F_{\beta}^{w}$)~\cite{margolin2014evaluate}, S-measure ($S_{\alpha}$)~\cite{fan2017structure}, and max E-measure ($E_{\xi}^{max}$)~\cite{fan2018enhanced}. mDice and mIoU are similarity measures at the regional level, while MAE and $F_{\beta}^{w}$ are utilized to evaluate the pixel-level accuracy. $S_{\alpha}$ and $E_{\xi}^{max}$ are used to analyze global-level similarity. The lower value is better for the MAE and the higher is better for others.

\begin{table}[t] \scriptsize
\caption{Quantitative results on Kvasir-SEG~\cite{jha2020kvasir} and CVC-ClinicDB~\cite{bernal2015wm} dataset. \textbf{Bold} indicates the best performance.}
	\label{tab:1Results}
 \centering 
\begin{tabular}{l|c|cccccc}
\toprule
\multicolumn{1}{l}{}  & \multicolumn{1}{l}{} & \multicolumn{6}{c}{\textbf{Kvasir-SEG}}                                                             \\ \cmidrule{3-8} 
method                & Year                 & mDice          & mIoU           & $F_{\beta}^{w}$            & $S_{\alpha}$             & $E_{\xi}^{max}$          & MAE            \\ \midrule
UNet~\cite{ronneberger2015u}                  & 2015                 & 0.818          & 0.746          & 0.794          & 0.858          & 0.893          & 0.055          \\
UNet++~\cite{zhou2018unet++}                & 2018                 & 0.821          & 0.744          & 0.808          & 0.862          & 0.909          & 0.048          \\
PraNet~\cite{fan2020pranet}                & 2020                 & 0.898          & 0.841          & 0.885          & 0.915          & 0.948          & 0.030          \\
ACSNet~\cite{zhang2020adaptive}                & 2020                 & 0.898          & 0.838          & 0.882          & 0.920          & 0.952          & 0.032          \\
SANet~\cite{wei2021shallow}                 & 2021                 & 0.904          & 0.847          & 0.892          & 0.915          & 0.953          & 0.028          \\
MSNet~\cite{zhao2021automatic}                 & 2021                 & 0.902          & 0.847          & 0.891          & 0.923          & 0.954          & 0.029          \\
SSFormer~\cite{wang2022stepwise}              & 2022                 & \textbf{0.925} & 0.878          & 0.921          & 0.931          & 0.969          & \textbf{0.017} \\
DCRNet~\cite{yin2022duplex}                & 2022                 & 0.846          & 0.772          & 0.807          & 0.882          & 0.917          & 0.053          \\
Polyp-sam~\cite{li2023polyp}                & 2023                 & 0.900         & 0.860         & -          & -         & -          & -          \\
\textbf{M$^{2}$ixNet(Ours)} & 2024                 & \textbf{0.929} & \textbf{0.881} & \textbf{0.929} & \textbf{0.935} & \textbf{0.971} & \textbf{0.014} \\ \midrule
\multicolumn{1}{l}{}  & \multicolumn{1}{l}{} & \multicolumn{6}{c}{\textbf{CVC-ClinicDB}}                                                           \\ \cmidrule{3-8} 
method                & Year                 & mDice          & mIoU           & $F_{\beta}^{w}$            & $S_{\alpha}$             & $E_{\xi}^{max}$          & MAE              \\ \midrule
UNet~\cite{ronneberger2015u}                  & 2015                 & 0.823          & 0.755          & 0.811          & 0.890          & 0.953          & 0.019          \\
UNet++~\cite{zhou2018unet++}               & 2018                 & 0.794          & 0.729          & 0.785          & 0.873          & 0.931          & 0.022          \\
PraNet~\cite{fan2020pranet}                & 2020                 & 0.899          & 0.849          & 0.896          & 0.937          & 0.979          & 0.009          \\
ACSNet~\cite{zhang2020adaptive}                 & 2020                 & 0.882          & 0.826          & 0.873          & 0.928          & 0.959          & 0.011          \\
SANet~\cite{wei2021shallow}                 & 2021                 & 0.916          & 0.859          & 0.909          & 0.940          & 0.976          & 0.012          \\
MSNet~\cite{zhao2021automatic}                  & 2021                 & 0.915          & 0.867          & 0.912          & 0.947          & 0.978          & 0.008          \\
SSFormer~\cite{wang2022stepwise}               & 2022                 & 0.916          & 0.873          & 0.924          & 0.937          & 0.971          & 0.007          \\
DCRNet~\cite{yin2022duplex}                & 2022                 & 0.869          & 0.800          & 0.832          & 0.919          & 0.968          & 0.023          \\
Polyp-sam~\cite{li2023polyp}                 & 2023                 & 0.920          & 0.870         & -         & -         & - & -         \\
\textbf{M$^{2}$ixNet (Ours)} & 2024                 & \textbf{0.941} & \textbf{0.891} & \textbf{0.934} & \textbf{0.955} & \textbf{0.989}          & \textbf{0.005} \\ \bottomrule
\end{tabular}
\end{table}

\begin{table}[h] \scriptsize
\caption{Quantitative results on ColonDB~\cite{tajbakhsh2015automated}, Endoscene~\cite{vazquez2017benchmark} and ETIS~\cite{silva2014toward}) dataset. \textbf{Bold} indicates the best performance.}
	\label{tab:2Results}
 \centering 
\begin{tabular}{l|c|cccccc}
\toprule
\multicolumn{1}{l}{}  & \multicolumn{1}{l}{} & \multicolumn{6}{c}{\textbf{CVC-ColonDB}}                                                                                                                               \\ \cmidrule{3-8} 
method                & Year                 & mDice                     & mIoU                     & $F_{\beta}^{w}$            & $S_{\alpha}$             & $E_{\xi}^{max}$          & MAE                        \\ \midrule
UNet~\cite{ronneberger2015u}                  & 2015                 & 0.504                     & 0.436                    & 0.491                   & 0.710                  & 0.781                     & 0.059                   \\
UNet++~\cite{zhou2018unet++}                & 2018                 & 0.481                     & 0.408                    & 0.467                   & 0.693                  & 0.763                     & 0.061                   \\
PraNet~\cite{fan2020pranet}                & 2020                 & 0.712                     & 0.640                    & 0.699                   & 0.820                  & 0.872                     & 0.043                   \\
ACSNet~\cite{zhang2020adaptive}                 & 2020                 & 0.716                     & 0.649                    & 0.697                   & 0.830                  & 0.851                     & 0.039                   \\
SANet~\cite{wei2021shallow}                 & 2021                 & 0.752                     & 0.669                    & 0.725                   & 0.837                  & 0.875                     & 0.043                   \\
MSNet~\cite{zhao2021automatic}                  & 2021                 & 0.747                     & 0.668                    & 0.733                   & 0.837                  & 0.883                     & 0.042                   \\
SSFormer~\cite{wang2022stepwise}               & 2022                 & 0.772                     & 0.697                    & 0.766                   & 0.844                  & 0.883                     & 0.036                   \\
DCRNet~\cite{yin2022duplex}                & 2022                 & 0.661                     & 0.576                    & 0.613                   & 0.767                  & 0.828                     & 0.110                   \\
Polyp-sam~\cite{li2023polyp}                 & 2023                 & \textbf{0.894}                    & 0.843                   & -                  & -                 & -                    & -                 \\
\textbf{M$^{2}$ixNet (Ours)} & 2024                 & 0.820            & \textbf{0.855}           & \textbf{0.799}          & \textbf{0.869}         & \textbf{0.935}            & \textbf{0.021}          \\ \midrule
\multicolumn{1}{l}{}  & \multicolumn{1}{l}{} & \multicolumn{6}{c}{\textbf{Endoscene} }                                                                                                                                \\ \cmidrule{3-8} 
method                & Year                 & \multicolumn{1}{l}{mDice} & \multicolumn{1}{l}{mIoU} & \multicolumn{1}{l}{$F_{\beta}^{w}$} & \multicolumn{1}{l}{$S_{\alpha}$} & \multicolumn{1}{l}{$E_{\xi}^{max}$} & \multicolumn{1}{l}{MAE} \\ \midrule
UNet~\cite{ronneberger2015u}                  & 2015                 & 0.710                     & 0.627                    & 0.684                   & 0.843                  & 0.875                     & 0.022                   \\
UNet++~\cite{zhou2018unet++}                 & 2018                 & 0.707                     & 0.624                    & 0.687                   & 0.839                  & 0.898                     & 0.018                   \\
PraNet~\cite{fan2020pranet}                & 2020                 & 0.871                     & 0.797                    & 0.843                   & 0.925                  & 0.972                     & 0.010                   \\
ACSNet~\cite{zhang2020adaptive}                 & 2020                 & 0.863                     & 0.787                    & 0.825                   & 0.923                  & 0.968                     & 0.013                   \\
SANet~\cite{wei2021shallow}                 & 2021                 & 0.888                     & 0.815                    & 0.859                   & 0.928                  & 0.972                     & 0.008                   \\
MSNet~\cite{zhao2021automatic}                  & 2021                 & 0.862                     & 0.796                    & 0.846                   & 0.927                  & 0.953                     & 0.010                   \\
SSFormer~\cite{wang2022stepwise}               & 2022                 & 0.887                     & 0.821                    & 0.869                   & 0.929                  & 0.962                     & 0.007         \\
DCRNet~\cite{yin2022duplex}                & 2022                 & 0.753                     & 0.670                    & 0.689                   & 0.854                  & 0.900                     & 0.025                   \\
Polyp-sam~\cite{li2023polyp}                 & 2023                 & \textbf{0.905}                     & \textbf{0.860}           & -                   & -                  & -                     & -                   \\
\textbf{M$^{2}$ixNet (Ours)} & 2024                 & 0.895            & \textbf{0.861}           & \textbf{0.879}          & \textbf{0.941}         & \textbf{0.978}            & \textbf{0.005}          \\ \midrule
\multicolumn{1}{l}{}  & \multicolumn{1}{l}{} & \multicolumn{6}{c}{\textbf{ETIS}}                                                                                                                                      \\ \cmidrule{3-8} 
method                & Year                 & mDice                     & mIoU                     & $F_{\beta}^{w}$                     & $S_{\alpha}$                     & $E_{\xi}^{max}$                     & MAE                     \\ \midrule
UNet~\cite{ronneberger2015u}                  & 2015                 & 0.398                     & 0.335                    & 0.366                   & 0.684                  & 0.740                     & 0.036                   \\
UNet++~\cite{zhou2018unet++}                & 2018                 & 0.401                     & 0.343                    & 0.390                   & 0.683                  & 0.776                     & 0.035                   \\
PraNet~\cite{fan2020pranet}                & 2020                 & 0.628                     & 0.567                    & 0.600                   & 0.794                  & 0.841                     & 0.031                   \\
ACSNet~\cite{zhang2020adaptive}                 & 2020                 & 0.578                     & 0.509                    & 0.530                   & 0.754                  & 0.764                     & 0.059                   \\
SANet~\cite{wei2021shallow}                 & 2021                 & 0.750                     & 0.654                    & 0.685                   & 0.849                  & 0.897                     & 0.015                   \\
MSNet~\cite{zhao2021automatic}                  & 2021                 & 0.720                     & 0.650                    & 0.675                   & 0.846                  & 0.881                     & 0.020                   \\
SSFormer~\cite{wang2022stepwise}               & 2022                 & 0.767                     & 0.698                    & 0.736                   & 0.863                  & 0.891                     & 0.016                   \\
DCRNet~\cite{yin2022duplex}                & 2022                 & 0.509                     & 0.432                    & 0.437                   & 0.714                  & 0.787                     & 0.053                   \\
Polyp-sam~\cite{li2023polyp}                 & 2023                 & \textbf{0.905}                     & 0.861                   & -                   & -                  & -                     & -          \\
\textbf{M$^{2}$ixNet (Ours)} & 2024                 & 0.891            & \textbf{0.866}           & \textbf{0.753}          & \textbf{0.882}         & \textbf{0.933}            & \textbf{0.009}          \\ \bottomrule
\end{tabular}
\end{table}

\begin{figure}[t]
	\begin{center}\scriptsize
		\tabcolsep 1pt
		\begin{tabular}{@{}c@{}}             
                \includegraphics[width = 1.0\textwidth]{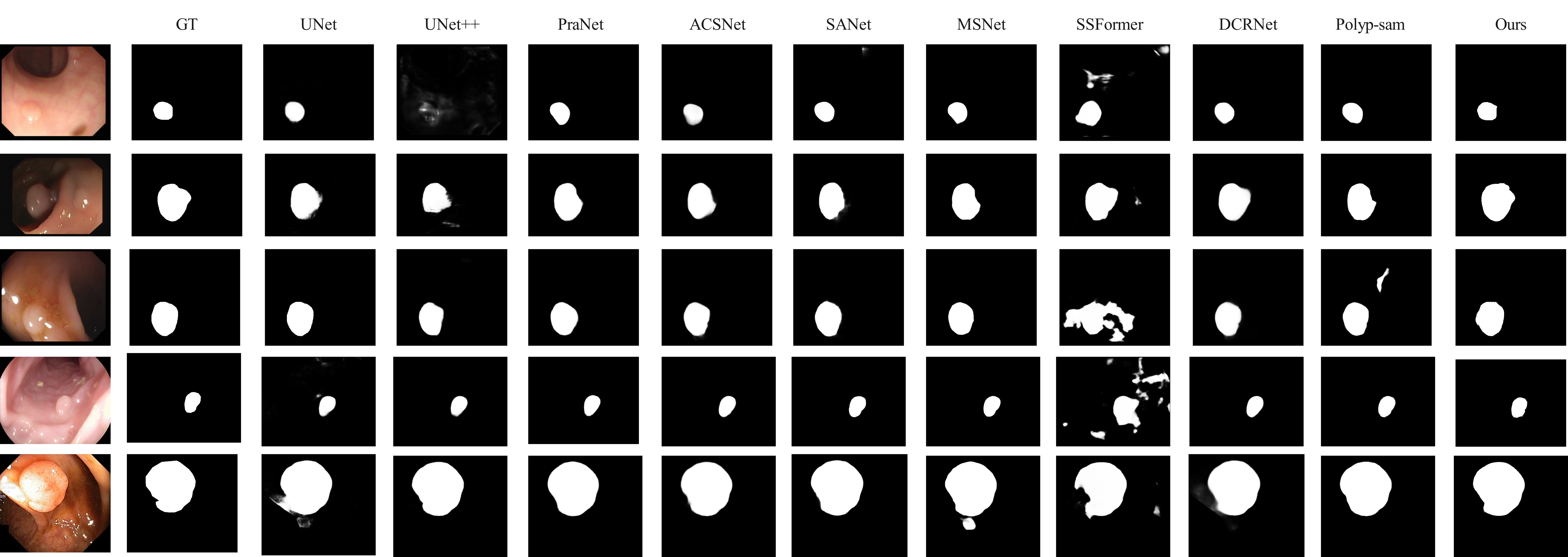}  			
		\end{tabular}
	\end{center}
 \vspace{-4mm}
 	\caption{Our method exhibits the best visual results.}
   \vspace{-2mm}
	\label{f5}
\end{figure}

\subsection{Experiments with State-of-the-art Methods}
We compare the proposed D$^{2}$LNet with previous state-of-the-art methods, including Polyp-sam~\cite{li2023polyp}, DCRNet~\cite{yin2022duplex}, SSFormer~\cite{wang2022stepwise}, MSNet~\cite{zhao2021automatic}, SANet~\cite{wei2021shallow}, ACSNet~\cite{zhang2020adaptive}, PraNet~\cite{fan2020pranet}, UNet++~\cite{zhou2018unet++}, UNet~\cite{ronneberger2015u}. For a fair comparison, we use their open-source codes to evaluate the same datasets or use the predictions provided by themselves. 

\noindent  \textbf{Quantitative Comparison.} As can be seen in Table~\ref{tab:1Results} and Table~\ref{tab:2Results}, M$^{2}$ixNet achieves the best scores across five datasets on almost all metrics.

\noindent  \textbf{Visual Comparison.}
Figure~\ref{f5} shows the visualization results of our model and the compared models. Our proposed model significantly outperforms other methods with better segmentation results,  which are the closest to the ground truth. We discover that our model possesses the following advantages: Our model can adapt to different lighting conditions, such as the overexposure observed in the third and sixth images in Figure~\ref{f5}. This is thanks to the role of the image-level prior that DAM brings into play.
%

\begin{table}[!htbp] \scriptsize
\caption{Ablation study for M$^{2}$ixNet on the Kvasir-SEG~\cite{jha2020kvasir} dataset.}
	\label{tab:5Results}
 \centering
\begin{tabular}{ccccccc}
\toprule
                  & mDice          & mIoU           & $F_{\beta}^{w}$            &  $S_{\alpha}$            & $E_{\xi}^{max}$          & MAE            \\ \midrule
w/o DAM & 0.874          & 0.840         & 0.882          & 0.899          & 0.930         & 0.049          \\
w/o multi-scale  & 0.911          & 0.876          & 0.920          & 0.910          & 0.969          & 0.020          \\
\textbf{M$^{2}$ixNet}   & \textbf{0.929} & \textbf{0.881} & \textbf{0.929} & \textbf{0.935} & \textbf{0.971} & \textbf{0.014} \\ \bottomrule
\end{tabular}
\vspace{-4mm}
\end{table}


\subsection{Ablation Study}
We conduct an ablation study to validate the effectiveness of the modules designed in our M$^{2}$ixNet. To validate the effectiveness of the DAM, we removed the role of depth map (w/o DAM). To analyze the effectiveness of the multi-scale input information, We use only the original polyp images as input (w/o multi-scale). Table~\ref{tab:5Results} demonstrates the performance of the model without the DAM and multi-scale drops sharply on Kvasir datasets. 


%

%

\section{Discussion and Conclusion}
We try to feed the depth maps generated by DAM as supplementary information to the UNet and UNet++ models, and on the Kvasir-SEG~\cite{jha2020kvasir} dataset, their performance improves by \textbf{5}\% and \textbf{8}\%, respectively.
Especially on extremely exposed datasets, DAM is also capable of generating high-quality depth maps to assist polyp segmentation models.

In this paper, we use the DAM-assisted polyp segmentation model to achieve accurate polyp segmentation.
Furthermore, our proposed M$^{2}$ixNet shows efficient segmentation.
%
%
%
Experimental results demonstrate the superiority of our method.


\bibliographystyle{splncs04}
\bibliography{mybib}

\end{document}